\documentclass[10pt,twocolumn,letterpaper]{article}

\usepackage[pagenumbers]{cvpr} % To force page numbers, e.g. for an arXiv version

\definecolor{cvprblue}{rgb}{0.21,0.49,0.74}
\usepackage[pagebackref,breaklinks,colorlinks,allcolors=cvprblue]{hyperref}

\usepackage{float}
\usepackage{listings}
\usepackage{bbm}
\usepackage{multirow}
\usepackage{booktabs}
\usepackage{colortbl}
\usepackage{transparent}
\usepackage{adjustbox}
\newcommand{\semitransp}[2][0.50]{{\transparent{#1}#2}}

\def\paperID{7056} % *** Enter the Paper ID here
\def\confName{CVPR}
\def\confYear{2025}

\title{Identifying and Mitigating Position Bias of Multi-image Vision-Language Models}
\author{
    Xinyu Tian\textsuperscript{\rm 1} \qquad
    Shu Zou\textsuperscript{\rm 1} \qquad
    Zhaoyuan Yang\textsuperscript{\rm 2} \qquad
    Jing Zhang\textsuperscript{\rm 1}
    \\
    \textsuperscript{\rm 1}Australian National University \quad \textsuperscript{\rm 2}GE Research
    \\
    {\tt\small \textsuperscript{\rm 1}firstname.lastname@anu.edu.au, \textsuperscript{\rm 2}firstname.lastname@ge.com}
}
\begin{document}
\maketitle

\begin{abstract}
\begin{figure*}[t] 
    \centering
    \includegraphics[width=1\linewidth]{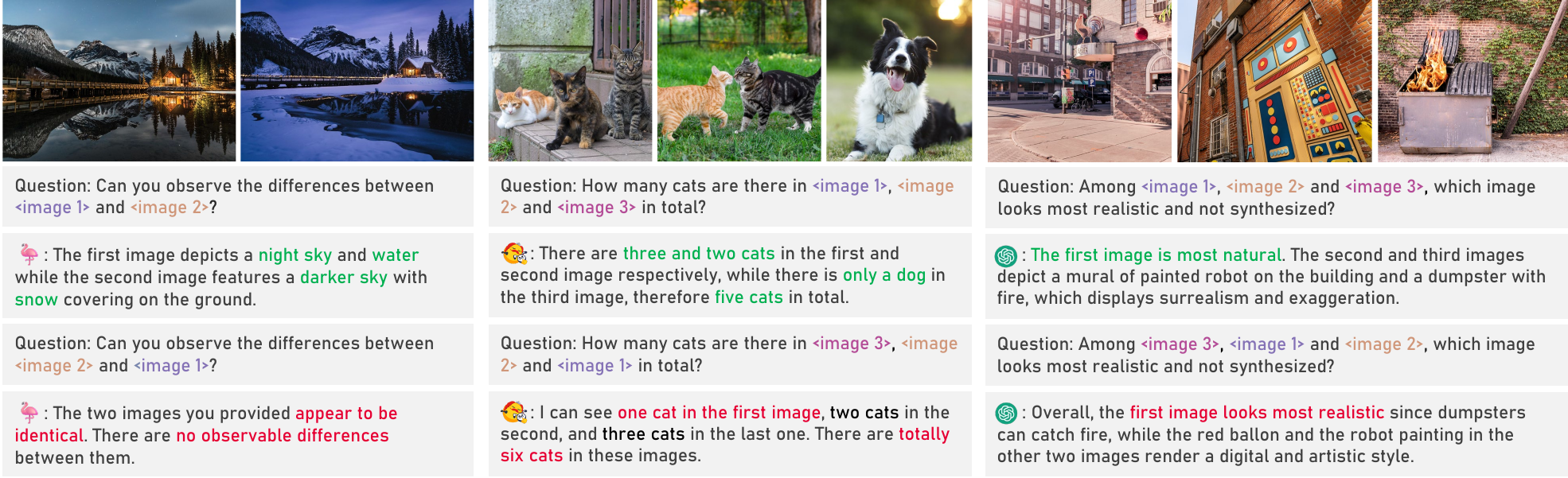}
    \caption{Examples where the predictions of LVLMs are  influenced by changes in the image positions \text{\footnotesize{(\texttt{Date}\:\:\texttt{accessed}: \texttt{Oct}\:\texttt{25},\:\texttt{2024})}}. The tasks, from left to right, are difference spotting~\citep{wang2024muirbench}, counting~\citep{jiang2024mantis} and forensic detection~\citep{fu2024blink}, respectively. The correct answers are highlighted in \textcolor[HTML]{00B050}{green}, while the incorrect ones are highlighted in \textcolor[HTML]{E60033}{red}. The prompts and outputs are simplified here for clarity and aesthetics.}
    \label{fig:failure_case}
    \vspace{-4mm}
\end{figure*}
The evolution of Large Vision-Language Models (LVLMs) has progressed from single to multi-image reasoning. Despite this advancement, our findings indicate that LVLMs struggle to robustly utilize information across multiple images, with predictions significantly affected by the alteration of image positions. To further explore this issue, we introduce Position-wise Question Answering (PQA), a meticulously designed task to quantify reasoning capabilities at each position. Our analysis reveals a pronounced position bias in LVLMs: open-source models excel in reasoning with images positioned later but underperform with those in the middle or at the beginning, while proprietary models show improved comprehension for images at the beginning and end but struggle with those in the middle. Motivated by this, we propose SoFt Attention (SoFA), a simple, training-free approach that mitigates this bias by employing linear interpolation between inter-image causal attention and bidirectional counterparts. Experimental results demonstrate that SoFA reduces position bias and enhances the reasoning performance of existing LVLMs. The code will be available \href{https://github.com/xytian1008/sofa}{\textcolor[HTML]{d12d8a}{https://github.com/xytian1008/sofa}}.
\end{abstract}    
\section{Introduction}
\label{sec:intro}
% Single-image to Multi-image LMMs
The advent of Large Vision-Language Models (LVLMs)~\citep{achiam2023gpt, liu2024visual, bai2023qwen, alayrac2022flamingo, laurenccon2024matters} has catalyzed the development of various text-grounded visual applications, including visual question answering~\citep{antol2015vqa, goyal2017making, lu2022learn}, code generation~\citep{suris2023vipergpt, lu2024chameleon, chen2024large}, and story telling~\citep{cohn2017picture, li2024seed}. Open-source LVLMs such as LLaVA~\citep{liu2024visual}, BLIP2~\citep{li2023blip}, and InternVL~\citep{chen2024internvl} have demonstrated competitive performance on single-image tasks like VQAv2~\citep{goyal2017making}, OKVQA~\citep{marino2019ok}, and MMMU~\citep{yue2024mmmu}. However, these LVLMs are limited in their capacity to analyze inter-image differences and relationships, as well as to process temporal information. Consequently, recent LVLMs, such as Flamingo~\citep{alayrac2022flamingo, awadalla2023openflamingo}, Idefics~\citep{laurenccon2024obelics, laurenccon2024matters, laurenccon2024building}, and Emu2~\citep{sun2024generative}, have incorporated image-text interleaved data during training to enable ability to tackle multi-image tasks~\citep{jiang2024mantis, wang2024muirbench, fu2024blink, suhr2018corpus}.

% Introduce position bias
Despite the increasing scale and performance of LVLMs, the interpretation of how they reason across multiple images remains a mystery. In fact, we might find some hints in the topic of natural language processing (NLP). Recent research in NLP has highlighted retrieval-augmented generation (RAG)~\citep{lewis2020retrieval, ram2023context, zhuang2023toolqa}, which aims to enable large language models (LLMs) to reason across multiple documents to answer questions. However,~\citep{liu2024lost} find that LLMs tend to effectively comprehend documents positioned at the beginning and end of the input, often neglecting the middle sections. Similarly, in LLM-as-a-judge~\citep{verga2024replacing, shi2024judging},~\citep{zheng2023judging} find that LLMs used to evaluate two model responses tend to favor the first one, regardless of content. This phenomenon, where the model's reasoning capability is influenced by the segment position, is referred to as position bias~\citep{liu2024lost, peysakhovich2023attention, wang2024eliminating}.

% Despite the increasing scale and performance of LVLMs, the understanding of how they reason across multiple images are rarely studied. In fact, this interpretation of multi-segment reasoning has been preliminarily studied in the field of natural language processing (NLP). Recent research in NLP has highlighted retrieval-augmented generation (RAG)~\citep{lewis2020retrieval, ram2023context, zhuang2023toolqa}, which aims to enable large language models (LLMs) to reason across multiple documents to answer questions. However,~\citep{liu2024lost} find that LLMs tend to effectively comprehend documents positioned at the beginning and end of the input, often neglecting the middle sections. Similarly, in LLM-as-a-judge~\citep{verga2024replacing, shi2024judging},~\citep{zheng2023judging} find that LLMs used to evaluate two model responses tend to favor the first one, regardless of content. This phenomenon, where the model's reasoning capability is influenced by the segment position, is referred to as position bias~\citep{liu2024lost, peysakhovich2023attention, wang2024eliminating}.

% Moitivate position bias in LVLMs
Motivated by the research above, a natural question arises: does position bias exist in LVLMs? Specifically, we seek to determine whether LVLMs can robustly make use of information across multiple images. To investigate this, we begin by evaluating LVLMs on established multi-image benchmarks~\citep{jiang2024mantis, wang2024muirbench, fu2024blink, suhr2018corpus}. We find that simply changing the order of presented images may significantly alter the predictions for most LVLMs, leading to considerable variation in benchmark accuracy. Fig.~\ref{fig:failure_case} provides examples of questions and model responses. For instance, in difference spotting~\citep{wang2024muirbench}, swapping two images causes OpenFlamingo~\citep{awadalla2023openflamingo} to miss distinctions, while in a simple counting task~\citep{jiang2024mantis}, reordering images leads Idefics2~\citep{laurenccon2024matters} to incorrect reasoning. This positional sensitivity is apparent even in GPT-4o~\citep{achiam2023gpt}, a leading proprietary model.

% Introduce Position-wise Question Answering
The above motivational study reveals an unfortunate fact: building upon LLMs, LVLMs may inherit the notorious position bias from their language component. This bias manifests as the model’s tendency to excel at reasoning visual information from certain positions, while exhibiting weaknesses at others, compromising robustness and reliability in predictions. To further investigate which positions display strong understanding and which reveal weaknesses, \ie, identifying the pattern of such position bias, we introduce Position-wise Question Answering (PQA). Unlike most multi-image tasks that assess holistic comprehension, PQA requires LVLMs to reason consistently across each image and produce position-wise responses. This approach enables us to measure position-wise accuracy and analyze the model's reasoning capacity across different positions. 

% Introduce the pattern of position bias
Our analysis reveals that most open-source models, \eg, OpenFlamingo~\citep{awadalla2023openflamingo} and Idefics2~\citep{laurenccon2024matters}, exhibit a pronounced recency bias, demonstrating stronger reasoning capabilities for images presented later while underperforming on those in the middle or at the beginning. In contrast, proprietary models like GPT-4o~\citep{achiam2023gpt} show enhanced understanding of images at the beginning and end, while neglecting those in the middle, a pattern consistent with previous findings in LLMs~\citep{liu2024lost}. This observation is further corroborated by Fig.~\ref{fig:failure_case}, where LVLMs tend to reason incorrectly about images positioned in poor-performing areas.

% Mitigating Position Bias
The existence of position bias significantly compromises model robustness and undermines accuracy across various multi-image tasks. Inspired by this, we aim to investigate the underlying mechanical causes and mitigate this issue. Specifically, we notice that in most LVLMs, the hidden states of image tokens are heavily influenced by their relative positions due to inter-image causal attention. In particular, image tokens are treated as a type of special text tokens where 1) images appearing later may interact with preceding ones; 2) images at the beginning are isolated, lacking access to global context. This autoregressive setup implicitly imposes a semantic order on the presented images, amplifying their positional information. To address this, we introduce SoFt Attention (SoFA), a training-free approach to mitigate position bias across various LVLMs. SoFA reduces the dependency of image hidden states on positional information by simply performing linear interpolation between inter-image causal attention and its bidirectional counterparts. Experimental results indicate that SoFA effectively equilibrates the reasoning capabilities of LVLMs across various positions, achieving average accuracy improvements of $2\sim3\%$ on existing multi-image benchmarks.

\noindent In summary, our contributions are as follows:
\begin{itemize}
    \item Despite the significant potential of LVLMs in multi-image applications, we find that position bias, a well-explored phenomenon in NLP, has emerged as a potential bottleneck in the development of LVLMs, substantially impacting model robustness and reliability.
    \item We further identify such position bias and discover that most open-source models demonstrate a pronounced recency bias, whereas proprietary models tend to struggle in middle positions.
    \item We introduce SoFA, a straightforward training-free approach designed to mitigate the position bias inherited from pre-trained LVLMs, and substantiate its effectiveness through experimental validation.
\end{itemize}
\begin{figure*}[t] 
    \centering
    \includegraphics[width=1\linewidth]{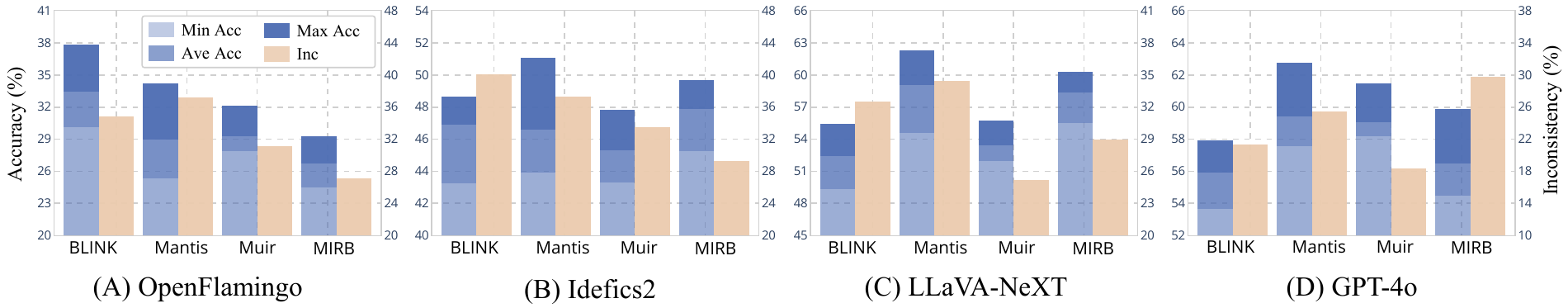}
    \caption{The results of multiple evaluations while shuffling image positions. We record the minimum, maximum, and average accuracy (left vertical axis), along with the prediction inconsistency between the best and worst-performing evaluations (right vertical axis).}
    \label{fig:motivation}
    \vspace{-4mm}
\end{figure*}
\section{Related Work}
\label{sec:related}
\noindent\textbf{Large Vision-Language Models.} 
Existing LVLMs are generally composed of a vision encoder~\citep{radford2021learning, zhai2023sigmoid}, an LLM~\citep{touvron2023llama, mosaicml2023introducing, chiang2023vicuna}, and a bridging projector~\citep{li2024llava, fuyu-8b, alayrac2022flamingo} that facilitates interaction between the two. Conventional LVLMs, pre-trained on image-text pairs~\citep{schuhmann2022laion} and fine-tuned through preference~\citep{liu2024llava, ouali2024clip} or instruction tuning~\citep{li2024llava, peng2023kosmos}, demonstrate substantial success in applications like dialogue~\citep{bai2023qwen, chen2024internvl}, question answering~\citep{wang2023cogvlm, zhu2023minigpt}, and domain-specific tasks~\citep{mh2024lvm, shentu2024cxr}. However, constrained by their training data, these models primarily support single-image reasoning during inference. Recently, the integration of image-text interleaved data~\citep{laurenccon2024obelics, zhu2024multimodal} has enabled multi-image reasoning capabilities, encompassing image comparison~\citep{wang2024muirbench, kazemi2024remi}, association~\citep{jiang2024mantis, fu2024blink}, temporal relationship analysis~\citep{li2024mvbench}, and emerging properties of multi-modal in-context learning~\citep{zhao2023mmicl, li2024configure}. Nonetheless, limited research has delved into how LVLMs perform reasoning across multiple images or the underlying mechanisms driving this process.

\noindent\textbf{Position Bias in LLMs.} 
Position bias, an inherent side effect of long-context input, has increasingly captured the attention of NLP researchers. The pioneering work on this issue, notably \citep{liu2024lost}, reveals that when LLMs process multiple documents, they often concentrate on those at the beginning and end while disregarding much of the information in between, \ie, a phenomenon termed ``lost in the middle". A related issue has also been identified in LLM-as-a-judge~\citep{verga2024replacing, shi2024judging}, where \citep{zheng2023judging} observe that LLMs tasked with evaluating multiple model responses tend to favor the first one, irrespective of its content. Recent studies have begun to explore methods to mitigate or circumvent position bias in LLMs~\citep{wang2024eliminating, yu2024mitigate, peysakhovich2023attention, ratner2022parallel}. However, to the best of our knowledge, few studies have examined the effects of position bias within multi-modal applications, specifically regarding its implications for LVLMs.

\noindent\textbf{Position Bias in LVLMs.} While systematic research on position bias in LVLMs is still lacking, recent studies have indirectly underscored the influence of positions on LVLMs through various findings. For instance, in multi-modal in-context learning, studies \citep{chen2023understanding} and \citep{baldassini2024makes} reveal that demonstration images placed at the beginning contribute minimally to LVLM performance. Additionally, in assessing long-context capabilities, \citep{wu2024visual} reports that LVLMs attain particularly high accuracy when key information appears at the end of the input sequence, with similar results observed in \citep{wang2024needle} and \citep{wang2024multimodal}. However, previous research has neither thoroughly analyzed this position bias nor proposed solutions to address it. In contrast, this study systematically investigates position bias in both open-source and proprietary models and introduces an effective mitigation strategy.

% Building upon these initial insights, this study systematically examines position bias in both open-source and proprietary models and introduces an effective mitigation strategy, \ie, SoFA.

\begin{figure*}[t] 
    \centering
    \includegraphics[width=1\linewidth]{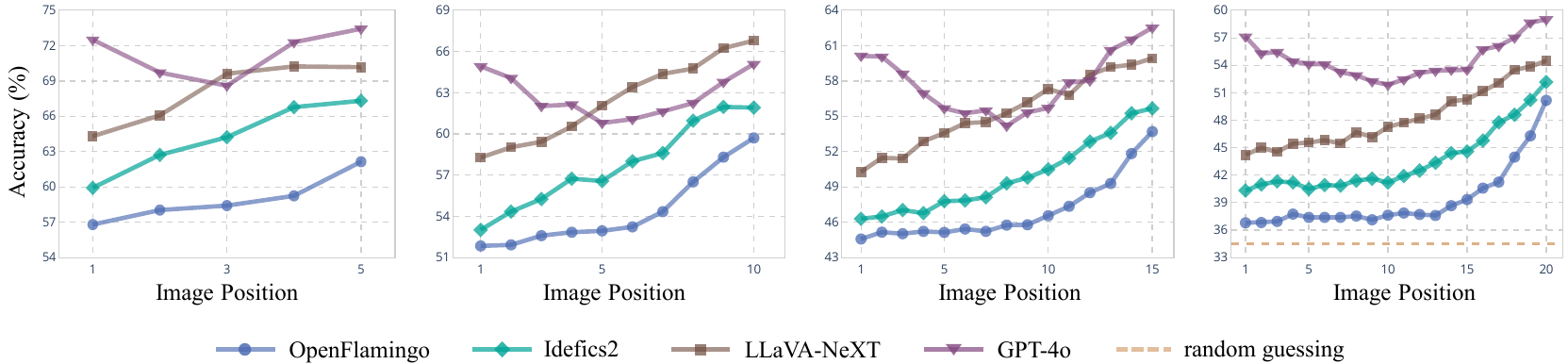}
    \caption{The results on the PQA task. We report position-wise accuracy on four scenarios where the number of images is 5, 10, 15 and 20, respectively. A higher accuracy signifies strong reasoning at the specified position, whereas lower reflects poor-performing areas.}
    \label{fig:pqa}
    \vspace{-4mm}
\end{figure*}
\section{Understanding Position Bias in LVLMs}
\label{sec:understand}
In this section, we investigate whether the position bias observed in LLMs extends to LVLMs. We start by empirically confirming the existence of position bias in LVLMs through a motivational study, revealing their poor robustness to image position changes (\hyperref[subsec:motivation]{§3.1}). Subsequently, we analyze the nature of this bias by evaluating and quantifying the reasoning capabilities of LVLMs regarding different positions (\hyperref[subsec:identify]{§3.2}). These insights contribute to a deeper understanding and interpretation of LVLMs' reasoning behavior and common failure cases.

% Subsequently, we analyze the nature of this bias by identify the reasoning capabilities of LVLMs at different positions.

\subsection{The Position Matters}
\label{subsec:motivation}
To assess the existence of position bias, we conduct a simple motivational experiment, as shown in Fig.~\ref{fig:motivation}.

\noindent\textbf{Experimental Setup}.  
We select four representative multi-image LVLMs: OpenFlamingo~\citep{awadalla2023openflamingo}, Idefics2~\citep{laurenccon2024matters}, LLaVA-NeXT-Interleave\footnote{For brevity, we refer to LLaVA-NeXT-Interleave as LLaVA-NeXT.}~\citep{li2024llava}, GPT-4o~\citep{achiam2023gpt}, and conduct evaluations across established multi-image benchmarks, including BLINK~\citep{fu2024blink}, Mantis-Eval~\citep{jiang2024mantis}, MuirBench~\citep{wang2024muirbench}, and MIRB~\citep{zhao2024benchmarking}. To ensure clarity, we select partial tasks from these benchmarks that feature position-agnostic questions, wherein the positions of the images do not affect the answers to the questions. See \hyperref[subsec:robustness]{§5.1} for the selected tasks and details. To explore the influence of image positions on the results, we perform multiple evaluations per benchmark, with the positions of the images shuffled in each iteration.

\noindent\textbf{Metric}. We record the minimum, maximum, and average accuracy across these evaluations, which directly reflects the influence of positions on the performance of LVLMs. Furthermore, we assess the prediction inconsistency between the best and worst-performing evaluations, quantifying the proportion of examples exhibiting conflicting predictions across the two evaluations by changing positions.

\noindent\textbf{Results.} 
As illustrated in Fig.~\ref{fig:motivation}, for most open-source LVLMs, the prediction inconsistency between the best and worst-performing evaluations approaches $30\%$, suggesting that merely altering the positions of the presented images may result in an amount of the model's predictions being changed. Such pronounced sensitivity to positional information results in highly unstable performance. For example, OpenFlamingo's accuracy on BLINK can vary significantly, reaching $38\%$ at its peak and dropping to $30\%$ at its lowest, indicating an accuracy span of up to $10\%$ with position changes. GPT-4o, the state-of-the-art proprietary model, although showcasing superior performance, also displays a lack of robustness to positional variations, leading to a prediction inconsistency of around $25\%$.

To qualitatively explore this issue, we present examples in Fig.~\ref{fig:failure_case} that illustrate prediction conflicts arising from alterations in image positions. For instance, when prompted to identify the differences between two images, OpenFlamingo fails to extract fine-grained distinctions upon swapping their positions. Similarly, in the task of object counting, Idefics2 mistakenly identifies a dog as a cat due to the repositioning of images. GPT-4o demonstrates a comparable issue, where rearranging the image options alters both its answer and the accompanying rationale. 
\subsection{Identifying Position Bias}
\label{subsec:identify}
The quantitative results and qualitative examples above imply biased reasoning across positions from LVLMs. To have a closer look at such position bias,  we aim to identify models' reasoning capabilities at different image positions.

However, existing tasks have been restricted to evaluating models' holistic comprehension of multiple images, lacking the means to systematically analyze their fine-grained understanding of each individual position. Specifically, these tasks provide coarse ground truth answers, such as option symbols or numeric values, but do not offer individual rationales for each image, making it challenging to quantify the model's reasoning across different positions. To address this gap, we introduce Position-wise Question Answering (PQA), a carefully designed task that assesses the position-wise reasoning capabilities of LVLMs. In PQA, each question prompts LVLMs to reason independently across all images, generating a distinct response for each. This approach 1) ensures task-level neutrality with respect to any particular position and 2) enables quantification of LVLMs' reasoning performance for each position.

Specifically, taking object counting as an example, we employ the following query template:
\begin{lstlisting}[breakatwhitespace=true]
I: Please provide one answer for each image in the form like [x, x, x, ...].
Q: Among <image 1>, ..., <image N>, how many cats can you find in these images?
A: [3, 2, 0, ...].
\end{lstlisting}
 Each query consists of an instruction (\texttt{I}) and a question (\texttt{Q}), where the former prompts the model to generate position-wise responses (\texttt{A}) following a specified output format. We gather responses from LVLMs as a structured list, with each element representing a particular position. This enables us to track position-wise accuracy, where a higher accuracy reflects stronger model capability at a given position, while lower indicates areas of relative weakness.

 \noindent\textbf{Construction of position-wise examples}. Since in PQA, each image is semantically independent of others, a PQA example essentially consists of several single-image instances that share the same question. This allows us to easily build position-wise examples from single-image benchmarks. Specifically,  we select VQAv2~\citep{goyal2017making}, a substantial single-image dataset that encompasses a diverse array of everyday topics and features both multiple-choice and open-ended questions. Initially, we identify and consolidate single-image examples with identical questions to create a question set, wherein each question corresponds to multiple images. Subsequently, we reformulate these questions into PQA format according to the above-mentioned template.

 \noindent\textbf{Experimental Setup.} 
We utilize the same LVLMs as previously mentioned and conduct evaluations on the proposed PQA task. Additionally, we also consider random guessing, where binary choice and numeric questions are answered randomly with yes or no, or with an integer value. We examine four scenarios with varying numbers of images: 5, 10, 15, and 20, collecting 1,000 examples for each scenario. Furthermore, to maintain fairness, we augment each example by shuffling the image positions four times, resulting in a total of 5,000 examples. In contrast to prior settings that evaluate holistic performance, we provide separate accuracy metrics for each position in this analysis.

\noindent\textbf{Results.} Fig.~\ref{fig:pqa} displays the results of LVLMs on the PQA task, from which we may derive the following insights:

\noindent 1) \textbf{Open-source LVLMs exhibit recency bias, whereas proprietary models struggle in middle positions}. For open-source models, a stronger reasoning capability is observed for images positioned at the end, while significant underperformance is noted in the middle and beginning. For example, with 10 images, the accuracy of OpenFlamingo varies by nearly $9\%$ between the far left and far right positions ($51.84\% \rightarrow 59.71\%$). In contrast, proprietary models, \ie, GPT-4o, reveal a typical U-shaped curve, highlighting pronounced weaknesses in the middle positions.

\noindent 2) \textbf{Position bias becomes more pronounced as the number of images increases}. For instance, with 20 images, the accuracy gap for OpenFlamingo between the two ends reaches $14\%$ ($36.79\% \rightarrow 50.17\%$). Notably, at this stage, OpenFlamingo seems to nearly lost its reasoning capability for most of the beginning images, achieving results only marginally better than random guessing ($34.49\%$).

\noindent 3) \textbf{Relationship to position bias in LLMs}. Previous research~\citep{liu2024lost} shows that LLMs often overlook information in middle positions when processing multiple documents, a phenomenon termed ``lost in the middle". Our findings suggest this effect persists for GPT in multi-image reasoning. However, for open-source models, the primacy position advantage appears to diminish, showing a marked decaying trend from the end to beginning. Whether the position bias in LVLMs is correlated with that of their LLM backbones remains an interesting question for further exploration.
\section{The Proposed Method}
\label{sec:method}
In this section, we first analyze the underlying causes of position bias (\hyperref[subsec:mechanism]{§4.1}). Then we introduce SoFA, a straightforward, training-free approach to mitigate this issue (\hyperref[subsec:mitigate]{§4.2}).
\subsection{Position Mechanism in LVLMs}
\label{subsec:mechanism}
To investigate the origins of position bias, we need to first determine what contributes to the positional information in LVLMs. In NLP,  it is generally believed that position embeddings~\citep{ratner2022parallel, su2024roformer} and causal attention~\citep{peysakhovich2023attention, wang2024eliminating} are the primary mechanisms enabling the sequential structure of input. Formally,
\begin{equation}
    \begin{aligned}
       &&\mathbf{Q}& = \mathcal{P}(\mathbf{XW_{q}}, \mathbf{pos}), \quad \mathbf{K} = \mathcal{P}(\mathbf{XW_{k}}, \mathbf{pos})\\ 
       &&\mathbf{H}& = \text{Softmax}(\mathbf{Q}\mathbf{K}^{\text{T}}/\sqrt{d})\odot \mathbbm{1}_{\text{causal}} \mathbf{V}
    \end{aligned}
\end{equation}
where $\mathbf{Q}$, $\mathbf{K}$, and $\mathbf{V}$ represent the queries, keys, and values, respectively. $\mathbf{X}$ refers to the input tokens, encompassing both image and text types, $\mathbf{pos}$ denotes the position embedding of $\mathbf{X}$, which is integrated into the attention via the position encoding function $\mathcal{P}$, $\mathbbm{1}_{\text{causal}}$ denotes the causal mask, which ensures autoregressive generation.

This reveals two key factors influencing LVLMs' positional awareness: 1) position embedding explicitly initializes a sequential value to each token, and 2) causal attention implicitly enhances this positional characteristic through unidirectional information access. This presents us with two distinct solutions. One approach is to directly reassign position embeddings, a widely used strategy in NLP~\citep{ratner2022parallel, chen2023fortify, he2024never}, as it fundamentally eliminates the influence of position on documents. However, this aggressive method is not suitable for LVLMs, since in many scenarios, such as video understanding and action recognition, we still expect images to retain positional information for temporal analysis. To clarify, our aim is to mitigate the imbalanced reasoning capabilities of LVLMs across different positions, rather than to eliminate their positional awareness.

\begin{figure}[t] 
    \centering
    \includegraphics[width=1\linewidth]{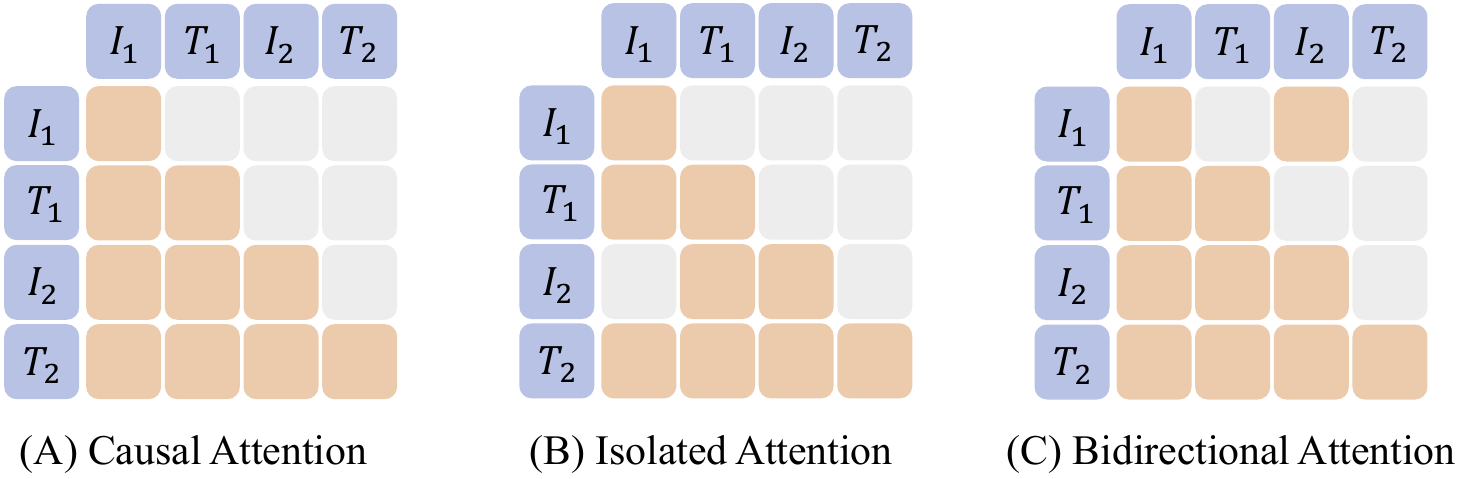}
    \caption{The three inter-image attention mechanisms, where $I_{1}$ and $I_{2}$ represent the tokens of two images, $T_{1}$ and $T_{2}$ represent the tokens of two text segments. Note that typically each image involves numerous tokens, \eg, 576 for LLaVA. Here for clarity we simplify them to a single token. In (A), images interact in a unidirectional manner, allowing $I_{2}$ to attend to $I_{1}$, while $I_{1}$ remains isolated. In (B), each image is isolated, indicating they can only attend to themselves. In (C), bidirectional interaction is enabled so that each image can attend to any other images. It is worth mentioning that we only alter the inter-image attention while preserving causal attention between the text segments.}
    \label{fig:attention}
    \vspace{-5mm}
\end{figure}
Therefore, in this paper, we primarily investigate the relationship between causal attention and position bias. We notice that, while images do not exhibit a distinct semantic order as text does, most LVLMs, \eg, LLaVA series, treat image and text tokens in a uniform autoregressive manner. As shown in Fig.~\ref{fig:attention} (A), analogous to texts, there exists inter-image causal attention which allows 1) later-positioned images to interact with those preceding them, while 2) earlier images remain isolated, lacking access to global contexts. From a mechanical perspective, this framework renders image hidden states highly dependent on their relative positions, thus implicitly amplifying their positional cues.

To verify our hypothesis, we compare the inter-image causal attention against following two alternative variants:

\noindent 1) \textbf{Isolated attention}. As illustrated in Fig.~\ref{fig:attention} (B), we disable interactions between images,  enforcing each image to attend solely to itself.

\noindent 2) \textbf{Bidirectional attention}. Rather than restricting to unidirectional information flow, we permit each image to attend to any other, as shown in Fig.~\ref{fig:attention} (C).

\begin{figure}[t] 
    \centering
    \includegraphics[width=1\linewidth]{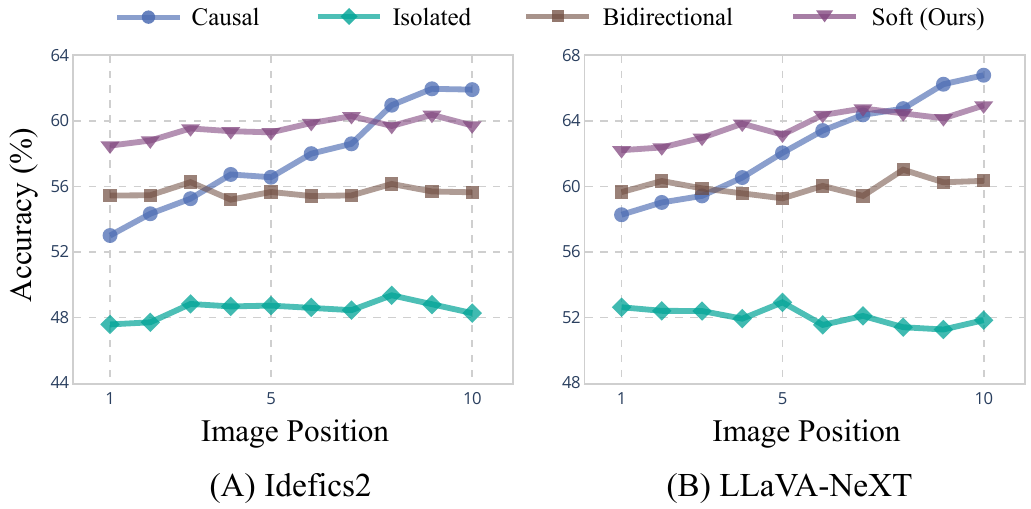}
    \caption{The position-wise accuracy of the three attention mechanisms on the PQA task, alongside our proposed SoFA method.}
    \label{fig:mask}
    \vspace{-4mm}
\end{figure}

We modify the default causal mask in the language decoder to its isolated and bidirectional counterparts and assess the impact of them on position bias in PQA, as shown in Fig.~\ref{fig:mask}. It is evident that altering causal attention to either isolated or bidirectional forms significantly alleviates position bias, resulting in more balanced accuracy across different positions. This empirically confirms that inter-image causal attention is the primary source of position bias. However, these modifications lead to varying degrees of performance degradation, with isolated attention causing the most substantial decline. This outcome is reasonable, as such modifications may substantially deviate from the training framework, leading to out-of-distribution hidden states.
\subsection{Mitigating Position Bias}
\label{subsec:mitigate}
The analysis above shows that inter-image causal attention is the main cause of position bias. Furthermore, it also reveals a trade-off between accuracy and robustness, a typical dilemma in machine learning. Building on this insight, we propose SoFt Attention (SoFA), a straightforward, training-free approach to mitigate position bias.  SoFA employs linear interpolation between inter-image causal attention and its bidirectional counterpart. Formally,
\begin{equation}
    \begin{aligned}
       &&\mathbf{H}_{\text{soft}}& = \text{Softmax}(\mathbf{Q}\mathbf{K}^{\text{T}}/\sqrt{d})\odot \mathbf{M}_{\text{soft}} \mathbf{V}\\ 
       &&\mathbf{M}_{\text{soft}}& = (1-\sigma) \mathbbm{1}_{\text{causal}} + \sigma \mathbbm{1}_{\text{bidirectional}}
    \end{aligned}
\end{equation}
where $\mathbbm{1}_{\text{causal}}$ and $\mathbbm{1}_{\text{bidirectional}}$ denote the two attention masks, with $\sigma$ controlling the extent of bidirectional part. Note that since we are only interested in the visual positional relation, we only modify the inter-image attention, while preserving causal attention between text segments.

\noindent\textbf{SoFA as a position smoother}. Intuitively, given that inter-image causal attention excessively emphasizes positional information, SoFA integrates partial bidirectional attention to smooth the influence of position on the models at minimal cost, achieving a balanced trade-off between reasoning performance and position robustness.

\noindent\textbf{SoFA as an interaction enhancer.} Practically, SoFA enhances the interaction between images, enriching the semantic information of earlier image features to enable more effective utilization by LVLMs. This aligns with the most cases where key information is typically distributed evenly across various input positions.

For each task, we utilize a small validation set to determine the optimal values for $\sigma$. Besides, to better align with the training framework, we deploy SoFA every two layers rather than at every layer, while preserving causal attention in the remaining layers. We present SoFA's results in the PQA task as shown in Fig.~\ref{fig:mask}. It is evident that SoFA significantly outperforms both isolated and bidirectional attention, while unlike causal attention in the default setting, SoFA demonstrates balanced performance and exhibits enhanced reasoning capability in most positions.

% SoFA essentially introduces a soft weighted mask between casual and bidirectional attention, thereby enhancing interactions between images and suppressing the dependency of hidden states on the images' relative positions.

% To implement SoFA, we need to determine the following two hyperparameters:

% \noindent 1) \textbf{The extent of bidirectional attention $\sigma$}. Intuitively, if $\sigma$ is too small, causal attention will dominate, resulting in an ineffective mitigation of position bias. Conversely, if $\sigma$ is too large, it will cause the attention mechanism to deviate from the autoregressive framework, compromising overall accuracy.

% \noindent 2) \textbf{The deployment layer interval $k$}. To align with the model's training framework, we deploy SoFA every $k$ layers in the language decoder rather than at every layer, while preserving causal attention in the remaining layers.
\section{Experiment and Results}
\label{sec:experiment}
\begin{table*}[t]
\footnotesize
\centering
\setlength{\tabcolsep}{1.4mm}
\setlength\heavyrulewidth{0.25ex}
\renewcommand{\arraystretch}{1.0}
\begin{tabular}{@{}lccccccccccccccccccc@{}}
\toprule
\multicolumn{1}{c}{\multirow{2}{*}{Method}} & \multicolumn{4}{c}{BLINK~\citep{fu2024blink}}&                                         \multicolumn{4}{c}{Mantis-Eval~\citep{jiang2024mantis}}                                  & \multicolumn{4}{c}{MuirBench~\citep{wang2024muirbench}}                                   & \multicolumn{4}{c}{MIRB~\citep{zhao2024benchmarking}}                                         \\ \cmidrule(r){2-5} \cmidrule(lr){6-9} \cmidrule(lr){10-13} \cmidrule(lr){14-17}
\multicolumn{1}{c}{}                        & Min            & Ave            & Max            & Inc           & Min            & Ave            & Max            & Inc           & Min            & Ave            & Max            & Inc           & Min            & Ave            & Max            & Inc           \\ \midrule
Idefics2~\citep{laurenccon2024matters}                                    & 43.26          & 46.93          & 48.68          & 41.55         & 43.93          & 46.62          & 51.10          & 38.57         & 43.28          & 45.30          & 47.81          & 34.46         & 45.26          & 47.88          & 49.68          & 29.94         \\
\rowcolor[HTML]{EDEDED}
Idefics2 + SoFA                             & 47.18          & 48.63          & 49.12          & 12.36         & 48.71          & 49.18          & 50.83          & \textbf{8.61} & 46.12          & 47.51          & 48.45          & 13.27         & 47.39          & 48.36          & 49.13          & 6.99          \\ \midrule
InternVL2~\citep{chen2024internvl}                                   & 38.81          & 40.45          & 43.74          & 30.18         & 45.36          & 46.92          & 49.51          & 25.24         & 49.28          & 52.33          & 56.10          & 38.65         & 41.84          & 44.38          & 46.36          & 29.60         \\
\rowcolor[HTML]{EDEDED}
InternVL2 + SoFA                            & 41.64          & 42.32          & 43.51          & \textbf{7.16} & 48.13          & 49.25          & 50.11          & 12.00         & \textbf{54.69} & \textbf{55.92} & \textbf{56.78} & \textbf{5.16} & 44.20          & 45.35          & 45.87          & \textbf{6.64} \\ \midrule
VILA~\citep{lin2024vila}                                       & 45.93          & 48.59          & 51.42          & 25.33         & 48.44          & 49.20          & 51.85          & 21.29         & 41.66          & 43.12          & 48.27          & 37.26         & 47.17          & 49.59          & 52.34          & 31.77         \\
\rowcolor[HTML]{EDEDED}
VILA + SoFA                                 & 48.27          & 50.80          & 51.17          & 10.68         & 49.29          & 51.60          & 52.68          & 12.16         & 45.76          & 46.52          & 47.13          & 7.92          & 48.22          & 51.43          & 51.95          & 17.46         \\ \midrule
Mantis~\citep{jiang2024mantis}                                      & 48.34          & 49.24          & 51.52          & 28.35         & 56.40          & 58.38          & 63.42          & 32.12         & 47.67          & 48.94          & 51.15          & 27.45         & 53.11          & 55.71          & 57.42          & 25.79         \\
\rowcolor[HTML]{EDEDED}
Mantis + SoFA                               & 49.22          & 50.87          & 52.34          & 16.23         & \textbf{60.48} & \textbf{62.21} & \textbf{64.68} & 14.30         & 49.88          & 50.26          & 50.79          & 5.61          & 54.39          & 56.34          & 56.93          & 8.35          \\ \midrule
LLaVA-NeXT~\citep{li2024llava}                                  & 49.34          & 52.40          & 55.43          & 31.89         & 54.57          & 59.10          & 62.29          & 33.75         & 51.96          & 53.45          & 55.78          & 24.92         & 55.49          & 58.40          & 60.28          & 28.56         \\
\rowcolor[HTML]{EDEDED}
LLaVA-NeXT + SoFA                           & \textbf{53.20} & \textbf{54.91} & \textbf{56.08} & 11.20         & 58.87          & 61.23          & 61.98          & 14.20         & 53.61          & 55.37          & 56.49          & 9.15          & \textbf{58.13} & \textbf{59.81} & \textbf{61.16} & 6.96          \\ \bottomrule
\end{tabular}
\caption{The evaluation on position-agnostic tasks with and without SoFA. Similar to \hyperref[subsec:motivation]{§3.1}, we perform multiple evaluations and record minimum, maximum and average accuracy. We also report prediction inconsistency between best and worst-performing evaluations.}
\label{tab:robustness}
\vspace{-4mm}
\end{table*}
In this section, we empirically demonstrate that SoFA 1) effectively mitigates the position bias inherent in LVLMs and 2) significantly improves their overall reasoning capability across various multi-image applications.

\noindent \textbf{Models}. 
We select a diverse set of open-source models, including Idefics2-8B~\citep{laurenccon2024matters}, InternVL2-8B~\citep{chen2024internvl}, VILA-13B~\citep{lin2024vila}, Mantis-8B~\citep{jiang2024mantis}, and LLaVA-NeXT-Interleave-7B~\citep{li2024llava}. 

\noindent \textbf{Implementation details}. For the input, we provide instructions that explicitly indicate the task type and output formats. For instance, for multiple-choice questions, we ask LVLMs to provide only the option symbol; for numeric questions, only the number; and for open-ended questions, the response is limited to short answers. For evaluation, to ensure fairness, we uniformly adopt GPT-4~\citep{achiam2023gpt}, providing with the question, ground truth answer, candidate prediction, and it determines whether the prediction is correct or not. For SoFA, we determine $\sigma$ using a 32-shot validation set per task and perform interpolation every two layers. Unless otherwise specified, all models are set to FP16 precision and utilize Flash Attention~\citep{dao2022flashattention}, with sub-image splitting~\citep{lin2023sphinx} disabled for fairness.
\subsection{Position Robustness}
\label{subsec:robustness}
\noindent\textbf{Benchmark}. 
We first evaluate whether SoFA may effectively mitigate position bias. To this end, similar to \hyperref[subsec:motivation]{§3.1}, we select position-agnostic tasks from existing multi-image benchmarks. Specifically, we exclude tasks where changing the order of images may alter the semantics of the question, rendering it unanswerable, \eg, video understanding. Here, we list some of the selected tasks: forensic detection and image jigsaw from BLINK~\citep{fu2024blink}, difference spotting and diagram understanding from MuirBench~\citep{wang2024muirbench}, visual analogy and attribute matching from MIRB~\citep{zhao2024benchmarking}. We include all examples from Mantis-Eval~\citep{jiang2024mantis} since the majority are position-agnostic. See the supplementary material for the details of more evaluated tasks. Our experimental setup and metrics are consistent with those described in \hyperref[subsec:motivation]{§3.1}.

\noindent \textbf{SoFA reduces the influence of positions on LVLMs}. As demonstrated in Table~\ref{tab:robustness}, the integration of SoFA within the models markedly reduces prediction inconsistency, leading to enhanced stability in performance. For instance, Idefics2's inconsistency on BLINK declines from $41.55\%$ to $12.36\%$, signifying that the responses for the majority of examples remain consistent despite variations in image positioning. This observation indicates that LVLMs exhibit more balanced reasoning capabilities across different positions, effectively mitigating the effects of position bias.

\noindent \textbf{SoFA eliminates the blind spots of LVLMs.} In addition to the enhanced robustness, we observe a marked improvement in the performance of LVLMs in worst-case scenarios, particularly regarding minimum and average accuracy, which both exhibit an increase of approximately $3\%$. This suggests that the model retains strong performance despite variations in the distribution of image positions, indicating that SoFA effectively addresses the deficiencies of LVLMs in positions where performance is typically suboptimal.
\subsection{Overall Results}
\noindent\textbf{Benchmark}. 
Given the observation that SoFA effectively mitigates position bias, we proceed with a comprehensive evaluation to determine whether SoFA can enhance the overall reasoning capabilities of LVLMs. Specifically, we include all tasks from the selected benchmarks~\citep{fu2024blink, jiang2024mantis, wang2024muirbench, zhao2024benchmarking}. In addition, we incorporate NLVR2~\citep{suhr2018corpus} and MVBench~\citep{li2024mvbench}, which exclusively assess the models' capability in image-pair comparison and video understanding, respectively. We directly utilize the default image positions from the benchmarks and record their accuracy.
% Previously, we seek to validate the effectiveness of SoFA in mitigating position bias by selecting position-agnostic tasks for evaluation. However, the presence of position bias can significantly impair the overall reasoning capabilities of LVLMs, thereby posing considerable risks across a range of multi-image tasks. In this section, we perform a comprehensive evaluation of SoFA's performance across all existing multi-image benchmarks. Specifically, for the previously identified four benchmarks~\citep{fu2024blink, jiang2024mantis, wang2024muirbench, zhao2024benchmarking}, we include all subtasks. Additionally, we consider NLVR2~\citep{suhr2018corpus} and MVBench~\citep{li2024mvbench}, which exclusively assess the models' capabilities in image-pair comparison and temporal analysis, respectively. We directly utilize the default image positions from the benchmarks and record their accuracy.

\begin{table}[t]
\footnotesize
\centering
\setlength{\tabcolsep}{0.6mm}
\setlength\heavyrulewidth{0.25ex}
\renewcommand{\arraystretch}{1.0}
\begin{tabular}{@{}lcccccc@{}}
\toprule
\multicolumn{1}{c}{Method}            & BLINK          & M-Eval    & Muir      & MIRB           & NLVR2          & MVBench        \\ \midrule
\semitransp{Random Guessing}          & \semitransp{39.27}         & \semitransp{25.31}          &  \semitransp{21.49}         & \semitransp{19.13}          & \semitransp{50.75}          & \semitransp{25.91}          \\ \midrule
Idefics2~\citep{laurenccon2024matters}          & 45.78          & 48.41          & 42.84          & 46.58          & 81.94          & 23.37          \\
\rowcolor[HTML]{EDEDED}
Idefics2 + SoFA   & 48.31          & 50.77          & 47.36          & 48.83          & 85.15          & 25.65          \\ \midrule
InternVL2~\citep{chen2024internvl}         & 38.95          & 50.30          & 54.53          & 42.66          & 85.56          & 29.31          \\
\rowcolor[HTML]{EDEDED}
InternVL2 + SoFA  & 43.26          & 51.11          & \textbf{57.14} & 46.19          & 88.19          & 32.77          \\ \midrule
VILA~\citep{lin2024vila}              & 49.79          & 52.64          & 48.73          & 50.36          & 78.82          & 25.81          \\
\rowcolor[HTML]{EDEDED}
VILA + SoFA       & 50.29          & 52.13          & 52.80          & 51.15          & 80.34          & 26.40          \\ \midrule
Mantis~\citep{jiang2024mantis}            & 51.41          & 56.85          & 44.28          & 54.59          & \textbf{90.26} & 41.59          \\
\rowcolor[HTML]{EDEDED}
Mantis + SoFA     & 53.19          & \textbf{59.23} & 47.64          & 55.68          & 89.98          & 44.40          \\ \midrule
LLaVA-NeXT~\citep{li2024llava}        & 53.34          & 50.83          & 48.22          & 57.15          & 87.28          & 54.26          \\
\rowcolor[HTML]{EDEDED}
LLaVA-NeXT + SoFA & \textbf{55.92} & 54.51          & 50.43          & \textbf{60.67} & 89.45          & \textbf{57.71} \\ \bottomrule
\end{tabular}
\caption{The overall evaluation accuracy of LVLMs on various multi-image benchmarks with and without SoFA.}
\label{tab:overall}
\vspace{-5mm}
\end{table}
\noindent\textbf{SoFA enhances reasoning capability of LVLMs.} In Table~\ref{tab:overall}, we observe that integrating SoFA leads to an average improvement of $2\sim 3\%$ in the performance of LVLMs across various benchmarks. For instance, Idefics2 shows a $2.25\%$ improvement on MIRB ($46.58\% \rightarrow 48.83\%$), while InternVL2 exhibits a $4.31\%$ increase on BLINK ($38.95\% \rightarrow 43.26\%$). This demonstrates that, despite the advancements in multi-image LVLMs, position bias poses a significant obstacle to their reasoning capability, whereas SoFA successfully addresses this limitation.
\subsection{Discussion}
\begin{figure}[t] 
    \centering
    \includegraphics[width=0.95\linewidth]{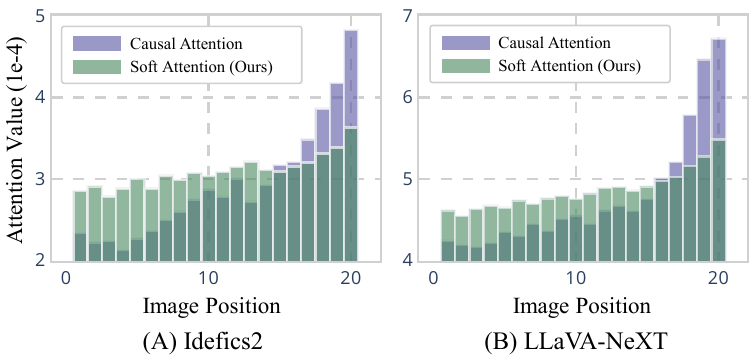}
    \caption{The attention distribution across positions on PQA.}
    \label{fig:attention_value}
    \vspace{-1mm}
\end{figure}

\noindent\textbf{Attention distribution visualization}. To gain a deeper understanding of SoFA's working mechanism, we visualize the attention distribution within LVLMs across different positions. Using PQA as an example, we derive the average attention value of the final layer's query token (whose hidden state is directly used for generation) for each image. As shown in Fig.~\ref{fig:attention_value}, under the default setting, the attention distribution is dominated by the end positions, indicating that the model overly focuses on a few images while neglecting most others. This observation aligns with our findings in \hyperref[subsec:identify]{§3.2}. The application of SoFA alters this trend by significantly smoothing the attention distribution, enabling the model to attend to images at all positions more evenly. This well explains how SoFA mitigates position bias.

\noindent\textbf{SoFA improves in-context capability of LVLMs.} Multi-modal in-context learning is an emerging property recently acquired by LVLMs, wherein a few demonstrations allow them to recognize task formats, thereby significantly improving accuracy. However, recent studies~\citep{li2024configure, chen2023understanding, baldassini2024makes} find that LVLMs tend to overlook demonstration images that appear earlier, leading to suboptimal performance. Here, we demonstrate that SoFA can improve the in-context learning capability of LVLMs by eliminating such blind spots. Following \citep{li2024configure}, We select three typical visual question answering benchmarks: VQAv2~\citep{goyal2017making}, VizWiz~\citep{gurari2018vizwiz}, and OK-VQA~\citep{marino2019ok}, and report the average accuracy of LVLMs across varying shot counts. As illustrated in Table~\ref{tab:icl}, the application of SoFA leads to a further enhancement in LVLM performance, with this effect becoming increasingly pronounced as the number of shots rises. For instance, in the case of VizWiz, accuracy improves from $41.56\%$ to $42.30\%$ with 4 shots, and from $45.35\%$ to $49.17\%$ with 16 shots. This indicates that position bias undermines the effectiveness of demonstrations, while SoFA mitigates this issue.

\begin{table}[t]
\footnotesize
\centering
\setlength{\tabcolsep}{0.6mm}
\setlength\heavyrulewidth{0.25ex}
\renewcommand{\arraystretch}{1.0}
\begin{adjustbox}{width=0.48\textwidth,center}
\begin{tabular}{@{}cccccccccccc@{}}
\toprule
\multirow{2}{*}{Method} & \multicolumn{3}{c}{VQAv2~\citep{goyal2017making}} & \multicolumn{3}{c}{VizWiz~\citep{gurari2018vizwiz}} & \multicolumn{3}{c}{OK-VQA~\citep{marino2019ok}} \\ 
\cmidrule(r){2-4} \cmidrule(r){5-7} \cmidrule(){8-10}
                        & 4-shot & 8-shot & 16-shot & 4-shot  & 8-shot & 16-shot & 4-shot  & 8-shot & 16-shot \\ \midrule
w/o SoFA                & 55.34  & 59.32  & 61.37   & 41.56   & 44.65  & 45.35   & 45.19   & 48.30  & 49.02   \\
\rowcolor[HTML]{EDEDED}
w/ SoFA                 & 56.13  & 61.30  & 64.85   & 42.30   & 46.98  & 49.17   & 45.93   & 49.87  & 52.36   \\ \bottomrule
\end{tabular}
\end{adjustbox}
\caption{The average accuracy of LVLMs on visual question answering using in-context learning with and without SoFA. }
\label{tab:icl}
\vspace{-1mm}
\end{table}
\begin{table}
\footnotesize
\centering
\setlength{\tabcolsep}{1.0mm}
\setlength\heavyrulewidth{0.25ex}
\renewcommand{\arraystretch}{1.0}
\begin{tabular}{@{}ccccccccc@{}}
\toprule
\# Images & 3     & 5     & 10    & 20    & 30    & 50    & 75                        & 100   \\ \midrule
w/o SoFA  & 82.16 & 81.27 & 77.18 & 72.36 & 68.50 & 60.80 & \multicolumn{1}{c}{55.13} & 49.19 \\
\rowcolor[HTML]{EDEDED}
w/ SoFA   & 83.90 & 82.33 & 78.92 & 74.30 & 70.67 & 63.27 & 59.38                     & 55.11 \\ \bottomrule
\end{tabular}
\caption{The average accuracy of LVLMs on Multi-modal Haystack~\citep{wang2024needle}, where images range from a few to a hundred.}
\label{tab:haystack}
\vspace{-5mm}
\end{table}
\noindent\textbf{SoFA exhibits advantages in long-context applications}. Long-context scenarios have recently become a focus, where a single inference requires processing a substantial number of images. This is particularly beneficial for applications such as many-shot in-context learning~\citep{jiang2024many} and multi-modal RAG~\citep{yasunaga2022retrieval}. Here, we show that SoFA can excel in these challenging scenarios. We select Visual Haystack~\citep{wu2024visual}, a benchmark designed to assess the long-context capabilities of LVLMs, and increase the input scale to encompass anywhere from a few images to a hundred. As shown in Table~\ref{tab:haystack}, the benefits of SoFA become increasingly evident as the number of images rises, achieving peak performance gains at 100 images ($49.19\% \rightarrow 55.11\%$). This indicates that an increase in context length exacerbates the model's position bias, while SoFA compensates this loss.

\begin{figure}[t] 
    \centering
    \includegraphics[width=0.95\linewidth]{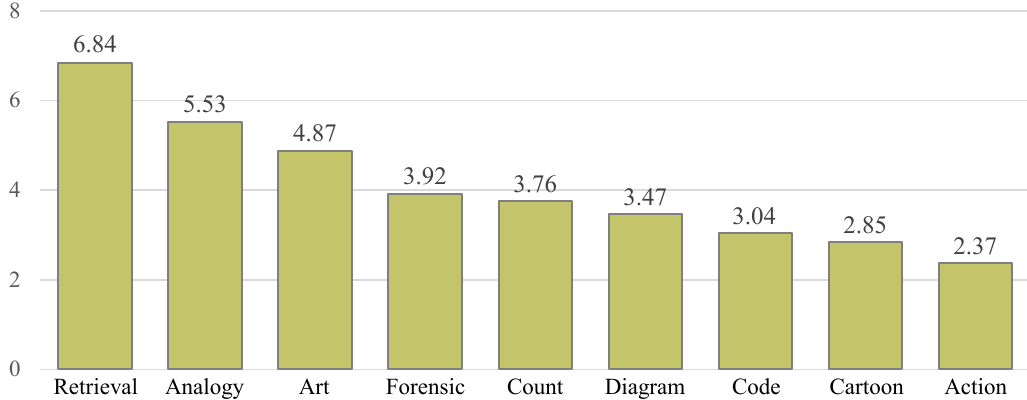}
    \caption{The performance gains of SoFA on different types of tasks. The results are averaged over selected models.}
    \label{fig:capability}
    \vspace{-2mm}
\end{figure}

\noindent\textbf{Contribution of SoFA to different tasks}. Extending beyond the holistic evaluation, we categorize the benchmarks into various tasks to identify which ones benefit the most from SoFA. As depicted in Fig.~\ref{fig:capability}, SoFA demonstrates the highest improvements in visual retrieval and analogy, with gains of $6.84\%$ and $5.53\%$, respectively. In these tasks, the first image acts as a reference, and the models need to select a matching scene or similar style from the remaining images. This necessitates that the model pays extra attention to and adequately understands the first image. Intuitively, due to the recency bias inherent in open-source models, they often overlook earlier information, resulting in suboptimal performance. Thus, the significant gains brought by SoFA are understandable. 

% \begin{table}[t]
% \footnotesize
% \centering
% \setlength{\tabcolsep}{0.6mm}
% \setlength\heavyrulewidth{0.25ex}
% \renewcommand{\arraystretch}{1.0}
% \begin{adjustbox}{width=0.48\textwidth,center}
% \begin{tabular}{@{}cccccccccc@{}}
% \toprule
% $\sigma$ & \rotbox{Retrieval} & \rotbox{Analogy} & \rotbox{Art} & \rotbox{Forensic} & \rotbox{Count} & \rotbox{Diagram} & \rotbox{Code}  & \rotbox{Cartoon} & \rotbox{Action}\\ \midrule
% 0.00   & 71.91          & 62.45              & 56.84               & 49.25           & 53.31             & 58.20             & 46.39          & 29.10               \\
% 0.25   & 73.04          & 63.50              & 58.98               & 51.17           & 55.83             & 59.81             & 48.76          & 32.25               \\
% 0.50   & 74.77          & 64.06              & 59.71               & 52.28           & 56.75             & 60.63             & \textbf{49.54} & \textbf{33.48}      \\
% 0.75   & 75.01          & 64.22              & 60.64               & \textbf{52.86}  & \textbf{57.49}    & \textbf{61.26}    & 49.43          & 32.69               \\
% 1.00   & \textbf{75.80} & \textbf{64.75}     & \textbf{61.07}      & 52.74           & 57.08             & 60.85             & 48.71          & 31.87               \\ \bottomrule
% \end{tabular}
% \end{adjustbox}
% \caption{The results on various subtasks varying the value of $\sigma$average accuracy of LVLMs on visual question answering using in-context learning with and without SoFA.}
% \label{tab:icl}
% \end{table}
\begin{table}[t]
\footnotesize
\centering
\setlength{\tabcolsep}{1.0mm}
\setlength\heavyrulewidth{0.25ex}
\renewcommand{\arraystretch}{1.0}
\begin{tabular}{@{}ccccccc@{}}
\toprule
$\sigma$ & BLINK          & Mantis-Eval    & MuirBench      & MIRB           & NLVR2          & MVBench        \\ \midrule
0.00  & 47.85          & 51.81          & 47.72          & 50.27          & 84.77          & 34.87          \\
0.25  & 49.22          & 52.70          & 49.35          & \textbf{52.34} & 85.41          & \textbf{37.28} \\
0.50  & \textbf{50.04} & 53.33          & \textbf{50.96} & 51.86          & 86.30          & 36.56          \\
0.75  & 49.62          & \textbf{53.74} & 48.24          & 51.14          & \textbf{86.83} & 35.07          \\
1.00  & 48.21          & 51.58          & 47.37          & 50.53          & 85.37          & 34.20          \\ \bottomrule
\end{tabular}
\caption{The results on six benchmarks varying the value of $\sigma$.}
\label{tab:sigma}
\vspace{-5mm}
\end{table}
\noindent\textbf{The interpretation of $\sigma$}. In Table~\ref{tab:sigma}, we conduct an ablation study on the effect of $\sigma$. It can be observed that the optimal $\sigma$ varies significantly across benchmarks, for instance, with a value of $0.75$ on NLVR2 and only $0.25$ on MVBench. Intuitively, MVBench inputs involve video frames, which have a strong dependence on image positions, and an excessively high $\sigma$ may compromise such positional information. In contrast, the examples in NLVR2 are image pairs, most of which are position-agnostic. This difference in task nature accounts for the variation in $\sigma$.

\section{Conclusion}
\label{sec:conclusion}
In this study, we introduce the concept of position bias, a well-known issue in NLP, into the realm of multi-modal applications. First, we reveal that LVLMs cannot robustly make use of information across multiple images, where merely altering the position of images may significantly change their predictions. Then we further identify this position bias, concluding that open-source models exhibit a strong recency bias, while proprietary models struggle with positions in the middle. Motivated by these findings, we propose SoFA, a simple, training-free approach that reduces the model's dependence on positional information through linear interpolation between inter-image causal attention and bidirectional counterparts. Experiments demonstrate that SoFA effectively mitigates position bias and significantly enhances the overall reasoning performance of existing models. 
\subsubsection*{Acknowledgement}
\label{sec:acknowledgement}
This research was, in part, funded by the U.S. Government – DARPA TIAMAT HR00112490421. The views and conclusions contained in this document are those of the authors and should not be interpreted as representing the official policies, either expressed or implied, of the U.S. Government.
{
    \small
    \bibliographystyle{ieeenat_fullname}
    \bibliography{main}
}

% \input{sec/X_suppl}
% {
%     \small
%     \bibliographystyle{ieeenat_fullname}
%     \bibliography{main}
% }
% WARNING: do not forget to delete the supplementary pages from your submission 

\end{document}